\documentclass[letterpaper]{article} 
\usepackage{aaai24}  
\usepackage{times}  
\usepackage{helvet}  
\usepackage{courier}  
\usepackage[hyphens]{url}  
\usepackage{graphicx} 
\usepackage{natbib}  
\usepackage{caption} 
\usepackage{algorithm}
\usepackage{algorithmic}

\usepackage{newfloat}
\usepackage{listings}

\usepackage{subcaption}
\usepackage{color, colortbl}
\usepackage{multirow}
\usepackage{graphicx}
\usepackage{amsmath}
\usepackage{colortbl}
\usepackage{bm}
\usepackage{amssymb}
\usepackage{dsfont}

\DeclareMathOperator*{\argmin}{arg\,min}

\DeclareCaptionStyle{ruled}{labelfont=normalfont,labelsep=colon,strut=off} 
\floatstyle{ruled}
\newfloat{listing}{tb}{lst}{}
\floatname{listing}{Listing}
\newcommand{\myPara}[1]{\vspace{.03in}\noindent\textbf{#1}}
\title{Coupled Confusion Correction: Learning from Crowds with Sparse Annotations}
\author{
Hansong Zhang\textsuperscript{\rm 1, 2}, Shikun Li\textsuperscript{\rm 1, 2}, Dan Zeng\textsuperscript{\rm 3}, Chenggang Yan\textsuperscript{\rm 4}, Shiming Ge\textsuperscript{\rm 1, 2}\thanks{Corresponding Author}
}

\affiliations{
    \textsuperscript{\rm 1}Institute of Information Engineering, Chinese Academy of Sciences, Beijing 100092, China\\
    \textsuperscript{\rm 2}School of Cyber Security, University of Chinese Academy of Sciences, Beijing 100049, China\\
    \textsuperscript{\rm 3}Department of Communication Engineering, Shanghai University, Shanghai 200040, China\\
    \textsuperscript{\rm 4}Hangzhou Dianzi University, Hangzhou 310018, China\\

    \{zhanghansong,lishikun,geshiming\}@iie.ac.cn, dzeng@shu.edu.cn, cgyan@hdu.edu.cn
}

\usepackage{bibentry}

\begin{document}

\maketitle

\begin{abstract}
As the size of the datasets getting larger, accurately annotating such datasets is becoming more impractical due to the expensiveness on both time and economy. Therefore, crowd-sourcing has been widely adopted to alleviate the cost of collecting labels, which also inevitably introduces label noise and eventually degrades the performance of the model. To learn from crowd-sourcing annotations, modeling the expertise of each annotator is a common but challenging paradigm, because the annotations collected by crowd-sourcing are usually highly-sparse. To alleviate this problem, we propose {\textbf{C}}oupled {\textbf{C}}onfusion {\textbf{C}}orrection (CCC), where two models are simultaneously trained to correct the confusion matrices learned by each other. Via bi-level optimization, the confusion matrices learned by one model can be corrected by the distilled data from the other. Moreover, we cluster the ``annotator groups'' who share similar expertise so that their confusion matrices could be corrected together. In this way, the expertise of the annotators, especially of those who provide seldom labels, could be better captured. Remarkably, we point out that the annotation sparsity not only means the average number of labels is low, but also there are always some annotators who provide very few labels, which is neglected by previous works when constructing synthetic crowd-sourcing annotations. Based on that, we propose to use Beta distribution to control the generation of the crowd-sourcing labels so that the synthetic annotations could be more consistent with the real-world ones. Extensive experiments are conducted on two types of synthetic datasets and three real-world datasets, the results of which demonstrate that CCC significantly outperforms state-of-the-art approaches. Source codes are available at: https://github.com/Hansong-Zhang/CCC.
\end{abstract}

\section{Introduction}
\begin{figure}[ht]
\centering
\includegraphics[width=0.46\textwidth]{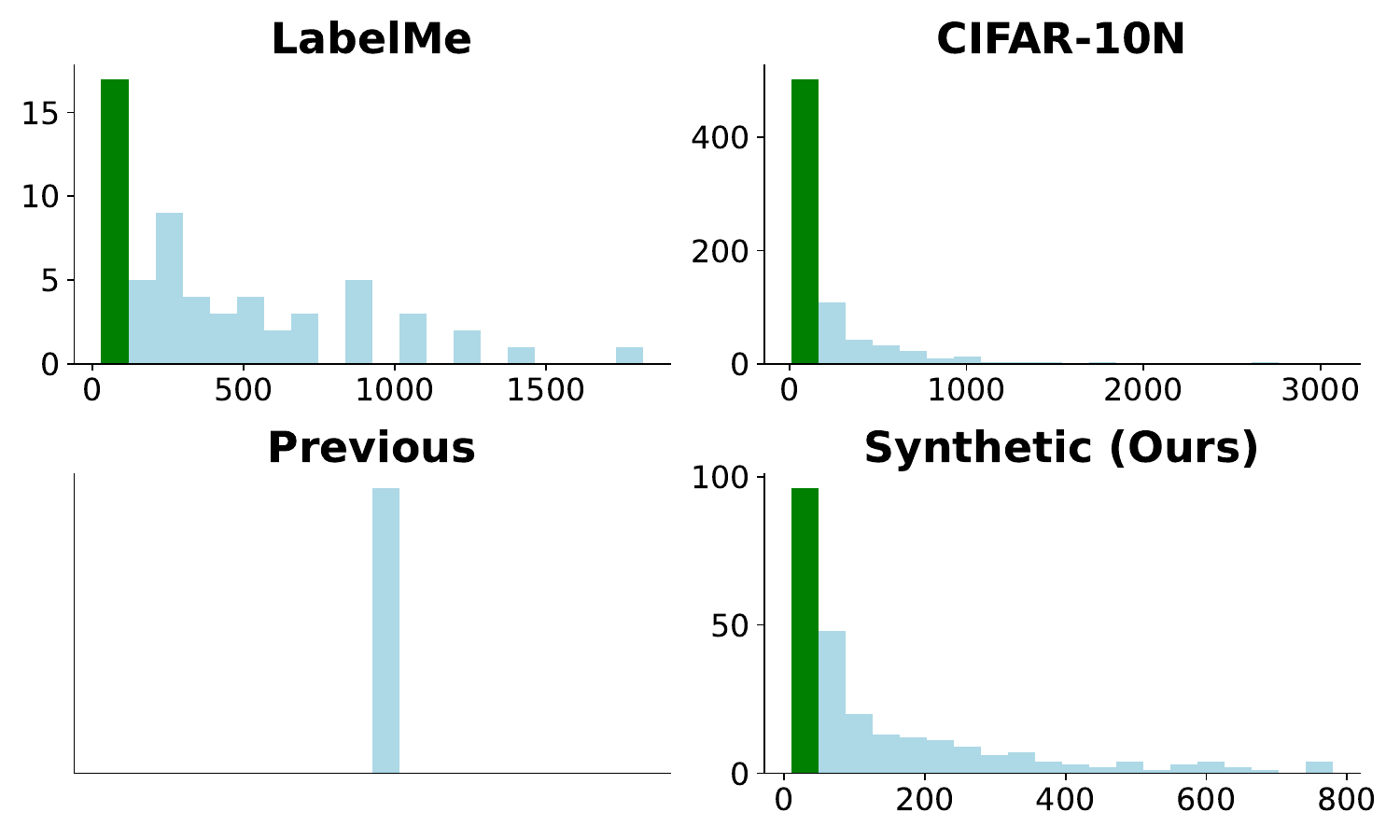}
\caption{The histogram of the number of labels provided by annotators. The green patch on the left denotes the annotators who only provide seldom labels.}
\label{label_number}
\end{figure}
The great success of Deep Neural Networks (DNNs) is much attributed to large-scale and accurately-labeled training datasets. However, collecting such high-quality annotations demands the annotators to be experts in their discipline and is often very time-consuming~\cite{song2022learning,han2018co,zhang2022knowledge}. For instance, the well-known {Microsoft COCO} dataset were constructed using 20K work hours in total~\cite{mscoco}. To improve the efficiency of construct such datasets, crowd-sourcing \cite{biel2012youtube,buecheler2010crowdsourcing,han2019millionaire,servajean2017crowdsourcing} provides an alternative way to collect annotations in a both time- and economic-efficient manner, where the labels of instances are provided by multiple annotators which could be non-expert. In the paradigm of crowd-sourcing, a large task is usually divided into several mini-tasks (one might overlap another), which will be further distributed to various annotators. On some crowd-sourcing platforms like \textit{Crowdflower}~\cite{crowdflower} and \textit{Amazon Mechanical Turk}, the annotators can get rewards when a mini-task is accomplished. Although crowd-sourcing is efficient in collecting large-scale annotations, it also inevitably introduces label noise~\cite{sheng2008get,cheapbutnoise_1,cheapbutnoise_2}, which is caused by the relatively-low expertise of some of the annotators. On the other hand, the DNNs can eventually memorize the noisy labels due to its great representative capacity~\cite{zhang2021understanding,memory-1,memory-2}. Therefore, increasing the generalization performance of DNNs trained with crowd-sourcing annotations is a challenging problem, which is widely termed as \textit{Learning from Crowds} (LFC).

In the discipline of LFC, previous solutions can be roughly divided into two pipelines: \textbf{(1)} \textit{Aggregate multiple noisy labels into a more reliable one before or during the training stage.} Among them, the simplest one is {Majority Voting}~\cite{majorityvote,han2019beyond,zhou2012ensemble}, where all the annotators are assumed to be equally reliable. Although {Majority Voting} provides a way to alleviate label noises, it is not always feasible to aggregate the labels equally, especially when the task is relatively hard and when most annotators are not reliable enough~\cite{majorityvote,han2019beyond,zhou2012ensemble}. Motivated by that, some advanced methods were designed to model the expertise of each annotator. For instance, {Weighted Majority Voting}~\cite{weightedmajor} endows the labels with a weight vector to alleviate the influence of the annotators with low expertise. \textbf{(2)} \textit{Training under the supervision of all annotations.} This pipeline of work usually take a set of confusion matrices as a measure of the expertise of annotators~\cite{crowdlayer, dawid1979maximum, conal, unionnet}. By simultaneously learning the confusion matrices and the classifier in an end-to-end manner, these methods have achieved more promising performance in LFC compared to the label-aggregating ones~\cite{crowdlayer, dawid1979maximum, conal, unionnet,Li2023TAIDTM}.

\textbf{Challenges.} Despite their improved performance, the modeling of the expertise of annotators suffers from the following two problems, which will eventually jeopardises the performance of the classifier. On the one hand, \textit{Annotation Sparsity} is a notorious property of the crowd-sourcing annotations~\cite{ccem}, which means that each annotator only labels a small subset of the whole dataset instead of all of it. On the other hand, the annotator-specific confusion parameters (ASCPs) (e.g. the weight vector or the confusion matrices in the aforementioned two pipelines) of each annotator are only learned based on its individual annotations. Therefore, if one annotator provides too few labels, it is obvious that the supervision of its ASCPs is insufficient and therefore its expertise can not be well-modeled (detailed analysis can be found in the Methodology Section). Note that such scenario is not hypothetical but quite ubiquitous. As shown in Figure~\ref{label_number}, in real-world crowd-sourcing datasets {LabelMe}~\cite{crowdlayer} and {CIFAR-10N}~\cite{cifarn}, a large number of annotators only provide seldom labels, even the authors of {CIFAR-10N} had enlarged the mini-tasks to alleviate annotation sparsity~\cite{cifarn}.

To mitigate the sparse annotation problem in LFC, in this paper we propose {\textbf{C}}oupled {\textbf{C}}onfusion {\textbf{C}}orrection (CCC). In our CCC, two simultaneously-trained classifiers are coupled in a way where the confusion matrices learned by one classifier can be corrected by a small \textbf{meta set} (a class-balanced set that contains the images and their pseudo labels which are likely to be clean) distilled from the other one. Note that the labels in the meta set are actually from the crowd-sourcing labels. As the meta set has dropped the information of annotator IDs, it can not be directly used to supervise the learning of ASCPs. To this end, we link the meta set and ASCPs via meta learning, in which way each annotator's ASCPs can be corrected by the meta set through bi-level optimization, even if the labels in meta set dose not come from him/her. Moreover, to further alleviate the annotation sparsity, we indicate and justify that there are many annotators who share similar expertise. Based on that founding, we use the {K-Means} method on the learned confusion matrices to cluster similar annotators so that their confusion matrices can be corrected together.

Our contributions can be summarized as follows:

\noindent$\bullet$ \textbf{A Meta-Learning-Based Method CCC to Mitigate the Sparse Annotation Problem.} In this work, we show that it is possible to learn one's ASCPs under the supervision of other annotations except for its own individual ones. Specifically, we link the meta set and ASCPs through meta learning paradigm, so that the ASCPs can be corrected by the meta set, enriching its supervision.
Moreover, we use {K-Means} to cluster annotators with similar expertise to further alleviate the annotation sparsity, correcting their ASCPs together.

\noindent$\bullet$ \textbf{Synthetic More Practical Crowd-Sourcing Annotation using Beta Distribution.} We point out a fact that is neglected by previous works: there exists some few-label annotators (annotators who only provide very few labels) in real-world crowd-sourcing datasets no matter how many the annotators are. To make our synthetic crowd-sourcing annotations more consistent with real-world ones, we propose to use Beta distribution to control the number of labels of each annotator. Previous works naively assumed that the amount of labels are evenly distributed, which may not accurately reflect real-world scenarios. Therefore, our proposed method is useful for future works on LFC when constructing synthetic annotations.

\noindent$\bullet$ \textbf{Extensive and Comprehensive Experiments.} We conduct experiments on various datasets, which include two types of synthetic crowd-sourcing datasets (independent and dependent cases) and three real-world datasets. The results demonstrate the competitiveness of our proposed method.
\section{Methodology}
\myPara{Notations and Preliminaries.}~We begin by fixing notations. Suppose that we are given a training dataset $\tilde{\mathcal{D}}=\{\bm{x}_i, \tilde{\bm{y}}_i\}_{i=1}^{N}$ which consists of $N$ instances labeled by $R$ annotators in total. Denote the data features as $\mathbf{x}=\{\bm{x}_i\}_{i=1}^N$ and the crowd-sourcing labels as $\mathbf{y}=\{\tilde{\bm{y}}_i\}_{i=1}^N$, where $\tilde{\bm{y}}_i=\left(\tilde{{y}}_i^{1}, \tilde{{y}}_i^{2},...,\tilde{{y}}_i^{R}\right)$. The $i$-th crowd-sourcing label $\tilde{\bm{y}}_i$ is a $R$-dimensional vector representing the labels provided by $R$ annotators. Note that it is usually the case that an annotator does not label all the data, so the crowd-sourcing labels may have missing ones. Let $C$ denotes the number of class categories, then the individual label $\tilde{y} \in \{1,2,...,C\}$. The goal of \textit{learning from crowds} is to train a classifier $f(\bm{x};\bm{\theta})$ parameterized by $ \bm{\theta} $ (we take DNN as the classifier in this paper, which takes the instance feature $ \bm{x} $ as input and outputs the possibility distribution over all classes) with training set $\tilde{\mathcal{D}}$, and to make it perform well on the unseen data. 
\subsection{Meta Learning Recap}\label{metarecap}
Next, we will briefly recap the preliminaries on \textit{Meta Learning}. In the meta-learning-based methods~\cite{mwnet,mlc,mlc_microsoft,famus, fast_reweighting}, an out-of-sample meta set 
$\mathcal{D}^{meta}=\{\bm{x}^{meta}_j, {y}_j^{meta}\}_{j=1}^M$ 
is assumed to be available, where $y_j^{meta}$ denotes the true label of the $j$-th meta data $\bm{x}^{meta}_j$ and $M (M \ll N)$ denotes the size of meta set. Note that the size of the meta set is much smaller than the training set, hence it is easy to collect in practice. The meta set can be either additionally provided~\cite{mwnet, mlc, mlc_microsoft} or distilled from the training set~\cite{famus, fast_reweighting}. Various hyper-parameters (e.g. instanced by the transition matrix~\cite{mlc}, label corrector net~\cite{mlc_microsoft} or multi-layer perceptron network~\cite{mwnet}) are called meta parameters and are updated via bi-level optimization, in which the meta set plays an important role in guiding the updating of such meta parameters. Intuitively, the performance of the current model on unseen data is measured by the meta set, the loss of the model on the meta set (termed as meta loss) is minimized in the outer loop of the bi-level optimization so that the gradient can flow back from the meta set to the meta parameters.

Typically, there are two sets of parameters to be optimized in such bi-level optimization, one is the aforementioned meta parameters, the other is the main parameters (mostly instanced by the parameters of the backbone model). Let $\bm{w}/\bm{\theta}$ denote the meta / main parameters, and $\mathcal{L}_i^{tra}(\bm{w},\bm{\theta})$ be the loss of the $i$-th instance in the training set $\tilde{\mathcal{D}}$. The optimal main parameters can be obtained by the training loss:
\begin{equation}
	\bm{\theta}^{\star}(\bm{w}) = \argmin_{\bm{\theta}} \frac{1}{N}\sum_{i=1}^{N} \mathcal{L}_i^{tra}(\bm{w}, \bm{\theta}).
 \label{inner}
\end{equation}
Then the meta parameters can be optimized by minimizing the meta loss:
\begin{equation}
	\bm{w}^{\star} = \argmin_{\bm{w}} \frac{1}{M}\sum_{j=1}^{M} \mathcal{L}_j^{meta}(\bm{\theta}^{\star}(\bm{w})),
 \label{outer}
\end{equation}
where the $\mathcal{L}_j^{meta}(\bm{\theta}^{\star}(\bm{w}))$ is the loss of the $j$-th instance in $\mathcal{D}^{meta}$. However, directly solving Eq.~(\ref{inner}) and Eq.~(\ref{outer}) is impractical, because the optimal $\bm{\theta}$ needs to be calculated whenever the meta parameters $\bm{w}$ is updated. Therefore, an online optimization method, as a trade-off between the feasibility and effectiveness, has been widely applied to the optimization of meta learning, which consists of three stages, including \textit{Virtual}, \textit{Meta}, and \textit{Actual} stage~\cite{mlc}. The optimization is performed in a mini-batch gradient descent manner, in $t$-th iteration, a training mini-batch $\tilde{\mathcal{M}} = \{\bm{x}_i, \tilde{\bm{y}}_i\}_{i=1}^n$ and a meta mini-batch $\mathcal{M}^{meta} = \{\bm{x}^{meta}_j, {y}^{meta}_j\}_{j=1}^m$ is fetched, where the $n$ and $m$ is the batch size of training and meta set respectively. Let $\eta_v, \eta_m, \eta_a$ denote the learning rates in the \textit{Virtual, Meta,} and \textit{Actual} stages respectively. In \textit{Virtual} Stage, the main parameters is ``virtually'' updated by:
\begin{equation}
	\hat{\bm{\theta}}_{t} = \bm{\theta}_{t-1} - \eta_v \frac{1}{n}\sum_{i=1}^{n} \nabla_{\bm{\theta}_{t-1}} \mathcal{L}_i^{tra}(\bm{w}_{t-1}, \bm{\theta}_{t-1}).
\end{equation}
This stage is named ``virtual'' because the true main parameters are kept unchanged, the above $\hat{\bm{\theta}}_{t}$ is fixed and only used to update the meta parameters in the following \textit{Meta} stage:

\begin{equation}
	\bm{w}_{t} = \bm{w}_{t-1} - \eta_m \frac{1}{m}\sum_{j=1}^{m} \nabla_{\bm{w}_{t-1}} \mathcal{L}_j^{meta}(\bm{w}_{t-1}, \hat{\bm{\theta}}_{t}).
\end{equation}
The \textit{Virtual} stage, together with the \textit{Meta} stage, can be deem as the outer loop of the bi-level optimization, which updates the meta parameters under the supervision of both meta set and training set~\cite{mlc}. The ``virtual'' main parameters $\hat{\bm{\theta}}_{t}$ is deleted as soon as the \textit{Meta} stage finished. Last, the $\bm{w}_{t}$ that comes from the outer loop will be fixed to update the main parameters in the \textit{Actual} stage:
\begin{equation}
	\bm{\theta}_{t} = \bm{\theta}_{t-1} - \eta_a \frac{1}{n}\sum_{i=1}^{n} \nabla_{\bm{\theta}_{t-1}} \mathcal{L}_i^{tra}(\bm{w}_{t}, \bm{\theta}_{t-1}).
\end{equation}
Unlike \textit{Virtual} stage, the main parameters in this stage is ``actually'' updated. After the above three stages, both the main and meta parameters are updated and ready for the $(t+1)$-th iteration. Note that the \textit{Actual} stage serves as the inner loop of the bi-level optimization. By alternating the outer and inner loop, the optimal main and meta parameters can be derived.

\subsection{How Annotation Sparsity Affects Training}\label{whysparseharmful}
\begin{figure*}[t]
\centering
\includegraphics[width=0.87\textwidth]{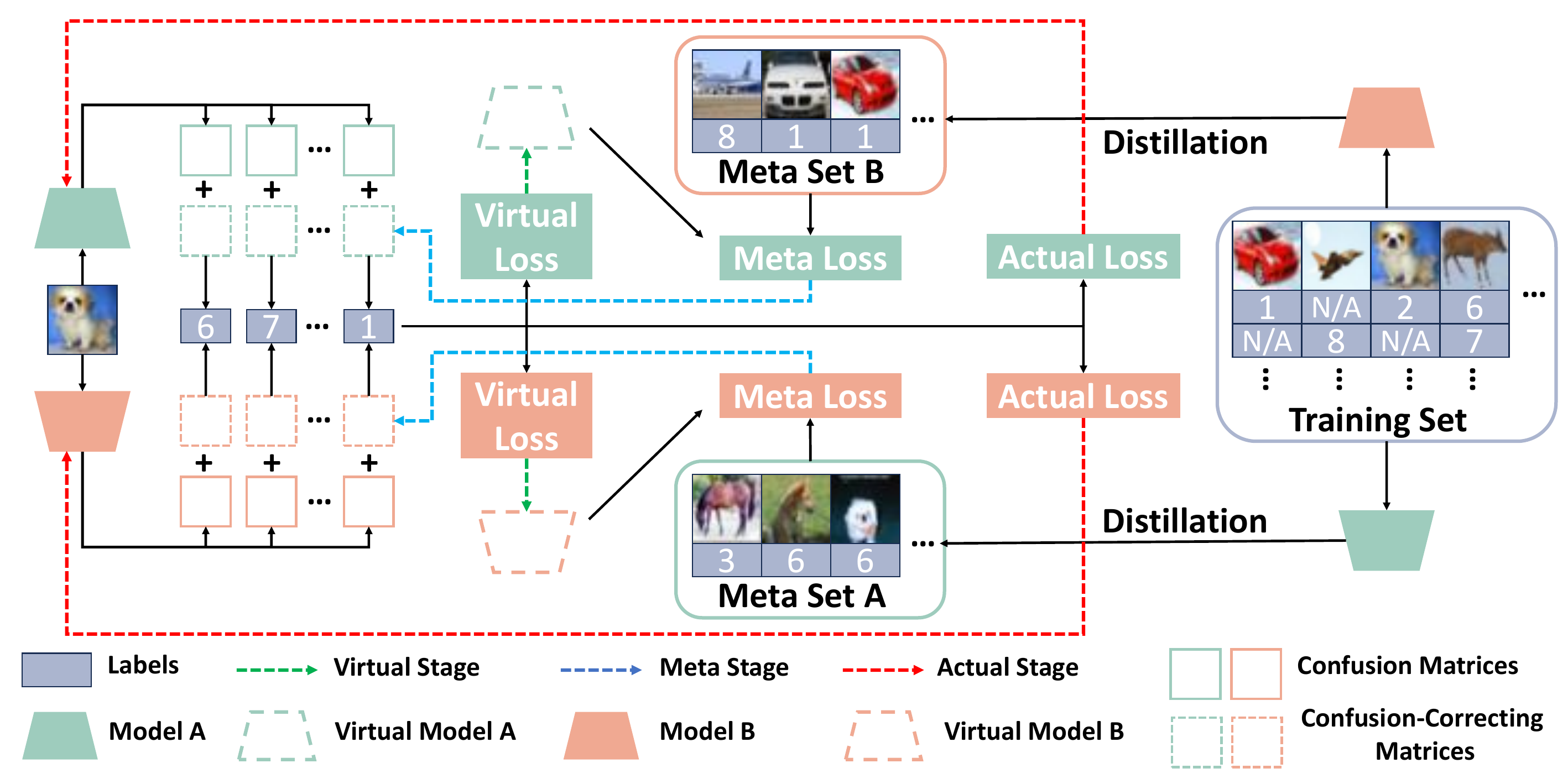}
\caption{The illustration of the our method. ``N/A'' means the label is missing.}
\label{framework}
\end{figure*}
In this section, we would like to show why the sparse annotation phenomenon is harmful, which motivates our work. For notation convenience, we use a matrix $\mathcal{A}_{(N\times R)}$ to indicate the annotation presence, i.e., $\mathcal{A}_{i,j}$ is {True} if the $i$-th instance is labeled by $j$-th annotator, and {False} otherwise. Without loss of generality, we hereby take {CrowdLayer}~\cite{crowdlayer} as an example, which uses $R$ confusion matrices $\{\bm{T}^r\}_{r=1}^R$ to model the annotator-specific expertise\footnote{The analyses also applies to the methods with other ASCPs.}. In {CrowdLayer}, the parameters of backbone DNN $\bm{\theta}$ and confusion matirces $\{\bm{T}^r\}_{r=1}^R$ are obtained by computing the following loss:
\begin{equation}
	\bm{\theta}^\star, \{\bm{T}^{r\star}\}_{r=1}^R = \argmin_{\bm{\theta},\{\bm{T}^r\}_{r=1}^R} \frac{1}{N}\sum_{i=1}^{N}\sum_{r=1}^{R} \mathds{1}[\mathcal{A}_{i,r}] \mathcal{L}_{i,r}(\bm{T}^r, \bm{\theta}),
 \label{crowdlayerloss}
\end{equation}
where $\mathds{1}[\cdot]$ denotes the indicator function, which outputs $1$ if the event is {True}, and $0$ otherwise. $\mathcal{L}_{i,r}(\bm{T}^r, \bm{\theta})=\ell (\bm{T}^r f(\bm{x}_i|\bm{\theta}) , \tilde{y}^r_i)$ stands for the loss of $i$-th instance and $r$-th annotator in {CrowdLayer}, where $\ell$ is a loss function such as widely-used cross entropy and $f$ is the softmax output of the classifier. For the $r_0$-th confusion matrix $\bm{T}^{r_0}$ (i.e. the ASCPs of $r_0$-th annotator), by modifying Eq.~(\ref{crowdlayerloss}), the gradient w.r.t. $\bm{T}^{r_0}$ (denoted by $\bm{g}^{r_0}$) can be derived as:
\begin{equation}
    \begin{aligned}
        \bm{g}^{r_0} &= \nabla_{\bm{T}^{r_0}} \frac{1}{N}\sum_{i=1}^{N}\sum_{r=1}^{R} \mathds{1}[\mathcal{A}_{i,r}] \mathcal{L}_{i,r}(\bm{T}^r, \bm{\theta})  \\
             &=\nabla_{\bm{T}^{r_0}} \frac{1}{N}\sum_{i=1}^{N} \mathds{1}[\mathcal{A}_{i,r_0}] \mathcal{L}_{i,r_0}(\bm{T}^{r_0}, \bm{\theta}),
    \end{aligned}
    \label{grad_CL}
\end{equation}
from which we notice that only the instances labeled by $r_0$-th annotator provide supervised information to its own confusion matrix, let alone the labels could be noisy. Therefore, if $r_0$-th annotator labels too few data, its confusion matrix would be poorly learned. It is worth noting that Figure~\ref{label_number} can be deemed as the histogram of $\sum_{i=1}^{N} \mathds{1}[\mathcal{A}_{i,r}]$, which highlights that a significant proportion of annotators contribute very few labels. This observation suggests that the issue mentioned above is prevalent in real-world scenarios.

\begin{figure}[tb]
\centering
\begin{subfigure}[b]{0.23\textwidth}
    \includegraphics[width=\textwidth]{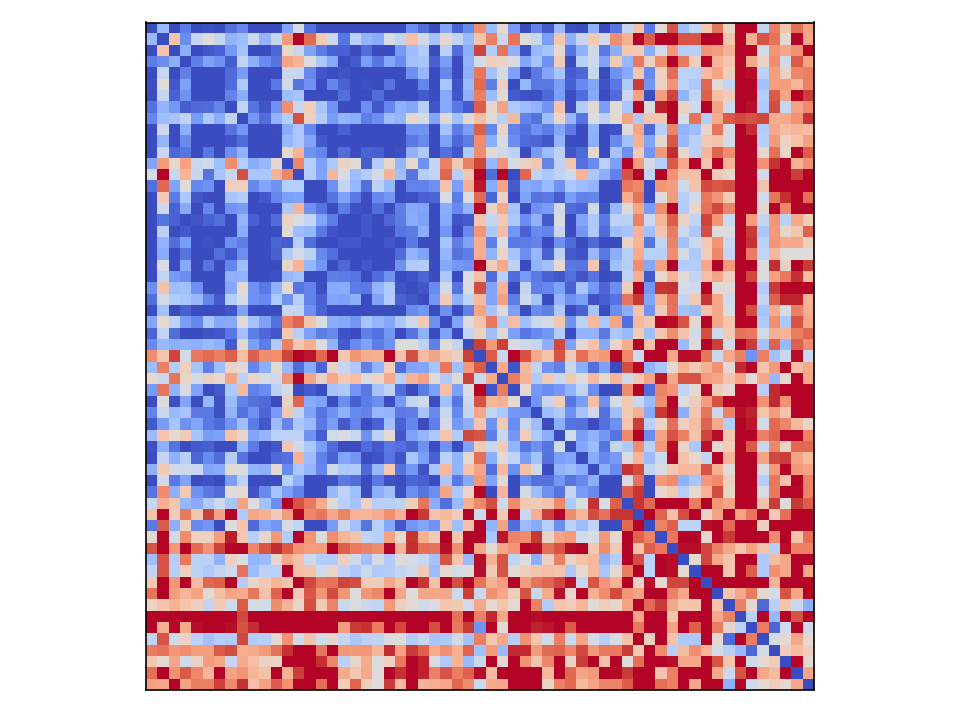}
    \caption{LabelMe $\#$1$\sim$59}
\end{subfigure}
\begin{subfigure}[b]{0.23\textwidth}
    \includegraphics[width=\textwidth]{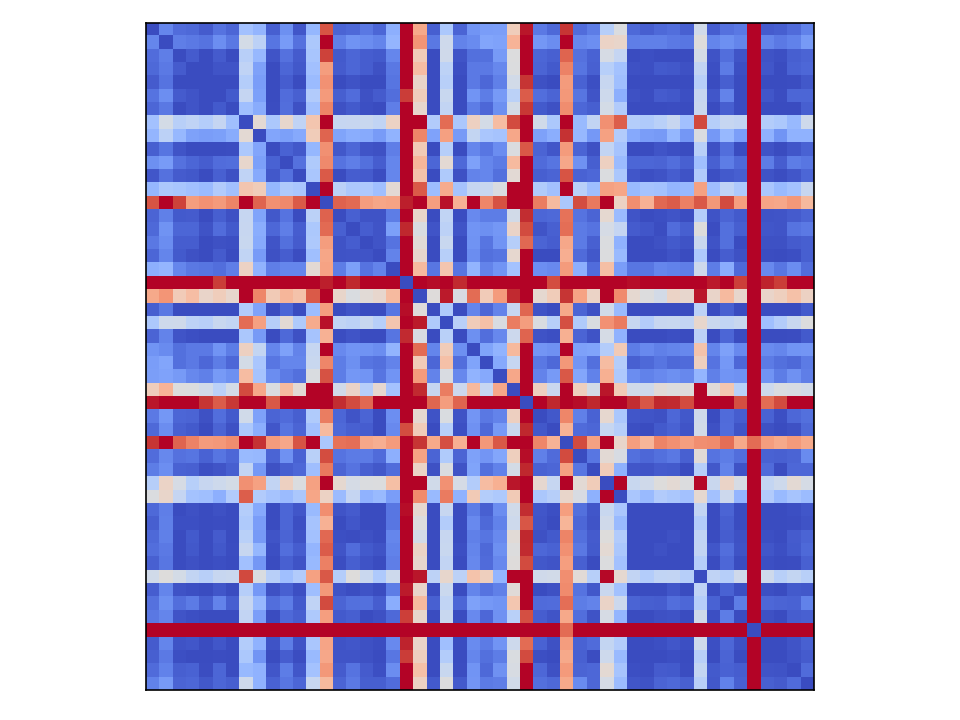}
    \caption{CIFAR-10N $\#$50$\sim$100}
\end{subfigure}
\caption{The distances of true confusion matrices of every two annotators in LabelMe and CIFAR-10N datasets. The $\#$ denotes the index of annotators. The more blue (red) the pixel is, the smaller (larger) value it is.}
\label{group}
\end{figure}
\subsection{Coupled Confusion Correction}\label{ourmethod}
Before diving into technical details, we first claim and justify an assumption that: \textit{In real-world crowd-sourcing datasets, there are various annotators who share similar expertise.}

We hereby take two common real-world crowd-sourcing datasets as examples, including {LabelMe} and {CIFAR-10N}~\cite{crowdlayer, cifarn, conal}. We calculate the true confusion matrices of each annotators $(CM^{r})_{(C\times C)}$ in such two datasets as follows:
\begin{equation*}
CM^{r}_{p,q} = 
\left\{
\begin{aligned}
    &0, \quad\quad\quad\quad\,\, \text{if} \sum_{i=1}^N \mathds{1}[\mathcal{A}_{i,r} \cap y_i=p] = 0\\
    &\frac{\sum_{i=1}^N \mathds{1}[\mathcal{A}_{i,r} \cap y_i=p \cap \tilde{y}_i^r=q]}{\sum_{i=1}^N \mathds{1}[\mathcal{A}_{i,r} \cap y_i=p]}, \quad \text{else} 
\end{aligned}
\right.
\end{equation*}
where $y_i$ denotes the true label of $i$-th instance. After getting the true confusion matrices, we calculate the distance between the true confusion matrices of every two annotators using mean-square-error.
Figure~\ref{group} illustrates the confusion distances between annotators in the {LabelMe} and {CIFAR-10N} datasets, where blue and red indicate small and large distances, respectively. As we can tell, there are many ``blue patches'' in the distance matrices, which suggests that the confusion matrices of annotators in such patch is similar to each other. Therefore, if their expertise can be considered together, the annotation sparsity can also be alleviated to some extent.

\myPara{Overview of Our Approach.}~In this paper, to alleviate the annotation sparsity, we propose {\textbf{C}}oupled {\textbf{C}}onfusion {\textbf{C}}orrection (CCC), where we simultaneously train two models and utilize the meta set distilled to correct the ASCPs learned by each other. Following~\cite{maxmig,conal,crowdlayer,aggnet}, we also adopt the confusion matrices to model the annotators' expertise. A set of confusion-correcting-matrices $\{\bm{V}^{r}\}_{r=1}^G$ are regarded as meta parameters to correct the confusion matrices of annotators $\{\bm{T}^{r}\}_{r=1}^R$, where $G$ is a hyper parameter indicating how many similar groups of annotators are. 
We run the K-Means~\cite{kmeans-1, kmeans-2} method on the learned confusion matrices $\{\bm{T}^{r}\}_{r=1}^R$ to obtain their group indexes at each epoch.
Unlike {CrowdLayer}~\cite{crowdlayer}, which only updates the confusion matrices based on individual annotations, CCC allows the confusion matrix of one annotator to be updated under the supervision of both his/her group's annotations and the meta set. This makes CCC more resilient to annotation sparsity, leading to improved classifier performance. In the following, we first explain why the supervised information can flow back from the meta set to the confusion-correcting-matrices. Then we present the distillation strategy of the meta set and the training objectives of different stages in our CCC. The framework of our method is illustrated in Figure~\ref{framework} and the pseudo code can be found in the Appendix.

\myPara{Correcting Confusion Matrices by Meta Set.}~In our CCC, the corrected confusion matrices are obtained through the following bi-level optimization:
\begin{equation}
    \begin{aligned}
            &\{\bm{V}^{r\star}\}_{r=1}^G = \argmin_{\{\bm{V}^r\}_{r=1}^G} \frac{1}{M}\sum_{j=1}^{M} \mathcal{L}_j^{meta}(\bm{\theta}^\star), \text{s.t.,}
\\
&\bm{\theta}^\star, \{\bm{T}^{r\star}\}_{r=1}^R = \argmin_{\bm{\theta}, \{\bm{T}^{r}\}_{r=1}^R} \frac{1}{N}\sum_{i=1}^{N}\sum_{r=1}^{R} \mathds{1}[\mathcal{A}_{i,r}] \mathcal{L}_{i,r}^{tra}(\bm{T}_{cor}^{r}, \bm{\theta})
    \end{aligned}
    \label{metaloss_}
\end{equation}
where $\bm{T}_{cor}^{r} = \bm{T}^{r} + \bm{V}^{g(r)}$ denotes the corrected confusion matrix of $r$-th annotator with $g(r)$ representing the index of group which the $r$-th annotator belongs to. Let $\ell$ denotes the cross entropy loss function, the meta loss here is then defined as $\mathcal{L}_j^{meta}(\bm{\theta}^\star)=\ell (f(\bm{x}^{meta}_j|\bm{\theta}^\star), y^{meta}_j)$, and the training loss is $\mathcal{L}_{i,r}^{tra}(\bm{T}_{cor}^{r},\bm{\theta})=\ell((\bm{T}^{r}+\bm{V}^{g(r)})f(\bm{x}_i|\bm{\theta}), \tilde{y}_i^r)$. Next, let us focus on the gradient w.r.t. the confusion-correcting-matrix of $r_0$-th annotator $\bm{V}^{g(r_0)}$ (denoted by $\bm{g}^{r_0}_{cor}$):

\begin{equation}
\scalebox{0.9}{$
\begin{aligned}
    &\bm{g}^{r_0}_{cor} = \nabla_{\bm{V}^{g(r_0)}} \frac{1}{M}\sum_{j=1}^{M} \mathcal{L}_j^{meta}(\bm{\theta}^\star) =\frac{1}{M}\sum_{j=1}^{M}\frac{\partial \mathcal{L}_j^{meta}}{\partial \bm{\theta}^\star}\cdot \frac{\partial \bm{\theta}^\star}{\partial \bm{V}^{g(r_0)}} \\
    &=\frac{1}{M}\sum_{j=1}^{M}\frac{\partial \mathcal{L}_j^{meta}}{\partial \bm{\theta}^\star}\cdot \frac{1}{N} \sum_{i=1}^{N}\sum_{r=1}^{R}\mathds{1}[\mathcal{A}_{i,r}] \nabla^2_{\bm{\theta}, \bm{V}^{g(r_0)}} \mathcal{L}_{i,r}^{tra} \\
    &=\frac{1}{MN}\sum_{j=1}^{M}\frac{\partial \mathcal{L}_j^{meta}}{\partial \bm{\theta}^\star}\cdot\sum_{i=1}^{N} \mathds{1}[\mathcal{A}_{i,r_0}] \nabla^2_{\bm{\theta}, \bm{V}^{g(r_0)}} \mathcal{L}_{i,r_0}^{tra}.
\end{aligned}
$}
\label{grad_MCC}
\end{equation}

Comparing Eq.~(\ref{grad_CL}) and Eq.~(\ref{grad_MCC}), we can see that every data in the meta set contribute to the overall gradient of the confusion-correcting-matrix of the $r_0$-th annotator. In this way, the learned confusion matrices can be corrected under the supervision of the meta set. Remarkably, because we cluster annotators with similar expertise, the confusion-correcting-matrices of two annotators ($r_0, r_1$) can be updated together as long as they belong to the same group $\left(g(r_0)=g(r_1)\right)$, which further hinders the influence of annotation sparsity.

\noindent \textbf{Distilling Meta Sets using Two Models.}~In most previous meta learning approaches~\cite{mwnet,mlc}, the meta set is demanded to be provided in addition to the training set, which may harms the fairness of the comparison to baselines. To this end, {FaMUS}~\cite{famus} and {FSR}~\cite{fast_reweighting} provided another way to construct meta set, where it can be distilled directly from the training set. However, their methods are constrained in the single-label scenario, hence we propose a new distilling strategy to match the crowd-sourcing annotations. The algorithm is quite simple to understand, as many sample-selection-based methods pointed out~\cite{memory-1, han2018co, dividemix}, the samples with smaller loss is more likely to be clean. Therefore, for each class $c$, we first fetch all the instances that are attached with the label $c$, then we select the ones with ${M}/{C}$ smallest losses. Nevertheless, it is worth noting that in Eq.~\ref{metaloss_}, if the losses of the instances that we select are already small, the supervised information that we obtain from the meta set may also be insufficient. Consequently, we simultaneously train two models to distill meta set for each other, in which way we can mitigate the confirmation bias (i.e., one model would accumulate its errors~\cite{dividemix}) to ensure the meta set to be sufficiently informative while maintaining clean.
\vspace{0.1cm}

\noindent \textbf{Training Objective.}~As stated above, in iteration $t$, we have a training mini-batch $\tilde{\mathcal{M}} = \{\bm{x}_i, \tilde{\bm{y}}_i\}_{i=1}^n$ and a meta mini-batch $\mathcal{M}^{meta} = \{\bm{x}^{meta}_j, {y}^{meta}_j\}_{j=1}^m$ for each model. The training objective and outcomes in three stages (\textit{virtual, meta, actual}) are shown as follows:
\begin{align}
	\text{Virtual}:&\, \frac{1}{n}\sum_{i=1}^{n}\sum_{r=1}^{R} \mathcal{L}^{tra}_{i,r}((\bm{T}^r_{t-1}+\bm{V}^{g(r)}_{t-1}), \bm{\theta}_{t-1}) \rightarrow \hat{\bm{\theta}}_{t} \label{virtual}\\
        \text{Meta}:&\, \frac{1}{m}\sum_{i=1}^{m} \mathcal{L}^{meta}_{j}(\hat{\bm{\theta}}_{t}) \rightarrow \{\bm{V}^r_{t}\}_{r=1}^G \label{meta}
\end{align}
Note that the confusion-correcting-matrices are initialized as zero matrices at the beginning of each iteration, after the above two stages (i.e. the outer loop), we have obtained the corrected confusion matrices $\bm{T}_{cor}^{r}|_{t-1} = \bm{T}^{r}_{t-1} + \bm{V}^{g(r)}_{t}$, which can be used in the following \textit{actual} stage (i.e. the inner loop):
\begin{align}
        \text{Actual}:&\, \frac{1}{n}\sum_{i=1}^{n}\sum_{r=1}^{R} \mathcal{L}^{tra}_{i,r}(\bm{T}_{cor}^{r}|_{t-1}, \bm{\theta}_{t-1}) \rightarrow \{\bm{T}^r_{t}\}_{r=1}^R, \bm{\theta}_{t} \label{actual}
\end{align}

\begin{table*}[tb]
    \centering
    \small
    \setlength\tabcolsep{4.5pt}
    \begin{tabular}{ll|cccc|cccc}
        \hline
        \multicolumn{2}{l|}{Dataset}  & \multicolumn{4}{c|}{CIFAR-10} & \multicolumn{4}{c}{Fashion-MNIST} \\
        \hline
        \hline
        \multicolumn{2}{l|}{Method / Case}         &  IND-\uppercase\expandafter{\romannumeral1} & IND-\uppercase\expandafter{\romannumeral2} & IND-\uppercase\expandafter{\romannumeral3} & IND-\uppercase\expandafter{\romannumeral4} &  IND-\uppercase\expandafter{\romannumeral1} & IND-\uppercase\expandafter{\romannumeral2} & IND-\uppercase\expandafter{\romannumeral3} & IND-\uppercase\expandafter{\romannumeral4}  \\
        \hline
        \multirow{2}{*}{MajorVote}   & Best & 55.08$\pm$0.38 & 47.85$\pm$2.20 & 54.27$\pm$1.91 & 66.94$\pm$1.01 & 84.89$\pm$2.24 & 79.94$\pm$2.30 & 76.72$\pm$1.00 & 89.03$\pm$0.75 \\
                                     & Last & 41.14$\pm$0.37 & 32.90$\pm$0.71 & 42.00$\pm$0.39 & 60.18$\pm$1.95 & 61.00$\pm$0.65 & 51.31$\pm$0.56 & 56.88$\pm$0.29 & 78.02$\pm$0.52 \\
        \hline
        \multirow{2}{*}{CrowdLayer}   & Best & 63.06$\pm$1.49 & 53.50$\pm$1.24 & 59.39$\pm$1.21 & 69.09$\pm$0.62 & 88.83$\pm$0.61 & 84.77$\pm$2.09 & 87.01$\pm$3.05 & 90.31$\pm$0.42 \\
                                      & Last & 54.78$\pm$0.49 & 43.71$\pm$0.62 & 53.17$\pm$0.34 & 67.51$\pm$0.91 & 75.89$\pm$0.19 & 66.39$\pm$0.21 & 66.67$\pm$0.47 & 84.67$\pm$0.20 \\
        \hline
        \multirow{2}{*}{DoctorNet}   & Best  & 69.63$\pm$0.86 & 63.43$\pm$1.46 & 69.77$\pm$2.02 & 76.25$\pm$0.47 & 90.54$\pm$0.23 & 88.78$\pm$0.68 & 88.29$\pm$1.20 & 90.96$\pm$0.32 \\
                                     & Last  & 66.80$\pm$1.12 & 57.95$\pm$1.09 & 65.40$\pm$0.98 & 74.60$\pm$0.45 & 83.01$\pm$0.13 & 76.07$\pm$0.49 & 75.77$\pm$0.20 & 86.78$\pm$0.19 \\
        \hline
        \multirow{2}{*}{Max-MIG}   & Best & 71.63$\pm$1.15 & 65.52$\pm$0.99 & 69.73$\pm$1.09 & 75.97$\pm$0.39 & 91.03$\pm$0.38 & 90.15$\pm$0.45 & 90.07$\pm$0.49 & 91.32$\pm$0.25 \\
                                   & Last & 66.47$\pm$0.48 & 56.63$\pm$0.46 & 61.68$\pm$0.63 & 73.19$\pm$0.59 & 82.16$\pm$0.32 & 73.33$\pm$0.39 & 73.53$\pm$0.47 & 86.16$\pm$0.37 \\
        \hline
        \multirow{2}{*}{CoNAL}      & Best & 67.28$\pm$1.24 & 59.27$\pm$1.53 & 65.93$\pm$0.15 & 78.16$\pm$0.27 & 89.95$\pm$0.30 & 87.62$\pm$0.88 & 89.13$\pm$0.44 & 91.72$\pm$0.14 \\
                                    & Last & 56.18$\pm$0.30 & 47.02$\pm$0.65 & 57.16$\pm$0.49 & 71.96$\pm$0.53 & 75.27$\pm$0.17 & 65.80$\pm$0.32 & 71.16$\pm$0.29 & 86.67$\pm$0.25 \\
        \hline
        \multirow{2}{*}{UnionNet}   & Best & 74.67$\pm$0.82 & 68.92$\pm$0.96 & 71.52$\pm$1.14 & 79.29$\pm$0.65 & 91.30$\pm$0.37 & 90.09$\pm$0.42 & 89.67$\pm$0.58 & 91.61$\pm$0.25 \\
                                    & Last & 74.44$\pm$0.60 & 63.41$\pm$0.75 & 67.00$\pm$0.42 & 78.78$\pm$0.74 & 88.22$\pm$0.19 & 78.29$\pm$0.29 & 79.09$\pm$0.54 & 90.56$\pm$0.21 \\
        \hline
        \multirow{4}{*}{CCC (Ours)} & Best & \textbf{80.16$\pm$0.23} & \textbf{75.33$\pm$0.43} & \textbf{78.28$\pm$0.25} & \textbf{83.06$\pm$0.26} & \textbf{92.59$\pm$0.01} & \textbf{91.93$\pm$0.18} & \textbf{92.50$\pm$0.13} & \textbf{93.23$\pm$0.06} \\
                                      & M.I. & $\uparrow$ 5.49 & $\uparrow$ 6.41 & $\uparrow$ 6.76 & $\uparrow$ 3.77 & $\uparrow$ 1.29 & $\uparrow$ 1.78 & $\uparrow$ 2.43 & $\uparrow$ 1.51\\\cline{3-10}
          & Last & \textbf{79.50$\pm$0.12} & \textbf{73.52$\pm$0.39} & \textbf{77.20$\pm$0.26} & \textbf{82.56$\pm$0.07} & \textbf{92.51$\pm$0.06} & \textbf{91.58$\pm$0.39} & \textbf{92.17$\pm$0.04} & \textbf{93.14$\pm$0.06}  \\
         & M.I. & $\uparrow$ 5.06 & $\uparrow$ 10.11 & $\uparrow$ 10.20 & $\uparrow$ 3.78 & $\uparrow$ 4.29 & $\uparrow$ 13.29 & $\uparrow$ 13.08 & $\uparrow$ 2.58 \\
         \hline
         \hline
        \multicolumn{2}{l|}{Method / Case}  &  COR-\uppercase\expandafter{\romannumeral1} & COR-\uppercase\expandafter{\romannumeral2} & COR-\uppercase\expandafter{\romannumeral3} & COR-\uppercase\expandafter{\romannumeral4} &  COR-\uppercase\expandafter{\romannumeral1} & COR-\uppercase\expandafter{\romannumeral2} & COR-\uppercase\expandafter{\romannumeral3} & COR-\uppercase\expandafter{\romannumeral4}  \\
        \hline
        \multirow{2}{*}{MajorVote}   & Best & 59.18$\pm$3.38 & 63.26$\pm$1.39 & 67.03$\pm$1.00 & 67.39$\pm$1.63 & 85.73$\pm$1.51 & 85.22$\pm$1.99 & 89.38$\pm$0.38 & 89.25$\pm$0.70 \\
                                     & Last & 43.57$\pm$0.61 & 50.36$\pm$0.76 & 59.04$\pm$0.62 & 57.97$\pm$1.07 & 60.83$\pm$0.57 & 66.11$\pm$0.83 & 79.69$\pm$0.77 & 75.68$\pm$0.46  \\
        \hline
        \multirow{2}{*}{CrowdLayer}   & Best & 62.51$\pm$1.12 & 67.45$\pm$1.03 & 69.24$\pm$0.69 & 70.07$\pm$0.40 & 88.49$\pm$0.44 & 89.87$\pm$0.17 & 90.19$\pm$0.28 & 89.88$\pm$0.23 \\
                                      & Last & 54.61$\pm$0.52 & 62.70$\pm$0.41 & 66.27$\pm$0.28 & 68.64$\pm$0.50 & 76.02$\pm$0.20 & 79.15$\pm$0.17 & 85.33$\pm$0.09 & 83.93$\pm$0.19 \\
        \hline
        \multirow{2}{*}{DoctorNet}   & Best & 67.30$\pm$1.28 & 74.01$\pm$0.62 & 73.87$\pm$0.48 & 77.52$\pm$0.49 & 89.29$\pm$0.37 & 90.54$\pm$0.32 & 90.36$\pm$0.21 & 90.86$\pm$0.21 \\
                                     & Last & 63.02$\pm$0.92 & 71.48$\pm$1.22 & 72.19$\pm$0.41 & 76.80$\pm$0.59 & 82.38$\pm$0.17 & 84.78$\pm$0.37 & 86.65$\pm$0.18 & 87.28$\pm$0.16 \\
        \hline
        \multirow{2}{*}{Max-MIG}   & Best & 71.69$\pm$0.68 & 74.55$\pm$0.56 & 76.83$\pm$0.30 & 75.81$\pm$0.68 & 90.27$\pm$0.07 & 91.26$\pm$0.17 & 91.38$\pm$0.16 & 91.17$\pm$0.28 \\
                                   & Last & 67.19$\pm$0.33 & 73.28$\pm$0.60 & 75.82$\pm$0.46 & 73.94$\pm$0.60 & 81.20$\pm$0.33 & 84.58$\pm$0.49 & 88.14$\pm$0.21 & 84.83$\pm$0.49  \\
        \hline
        \multirow{2}{*}{CoNAL}      & Best & 69.33$\pm$0.46 & 72.21$\pm$0.30 & 75.97$\pm$0.19 & 79.82$\pm$0.49 & 89.48$\pm$0.34 & 90.79$\pm$0.40 & 91.37$\pm$0.10 & 92.02$\pm$0.14 \\
                                    & Last & 58.24$\pm$0.63 & 64.10$\pm$0.27 & 68.42$\pm$0.65 & 73.82$\pm$0.83 & 75.49$\pm$0.30 & 78.36$\pm$0.28 & 85.93$\pm$0.40 & 88.47$\pm$0.47 \\
        \hline
        \multirow{2}{*}{UnionNet}   & Best & 74.46$\pm$0.43 & 78.13$\pm$0.43 & 80.02$\pm$0.51 & 80.99$\pm$0.66 & 90.83$\pm$0.40 & 91.52$\pm$0.14 & 91.94$\pm$0.22 & 92.31$\pm$0.22 \\
                                    & Last & 72.84$\pm$0.31 & 77.58$\pm$0.47 & 79.37$\pm$0.62 & 80.85$\pm$0.66 & 85.32$\pm$0.21 & 88.60$\pm$0.37 & 91.82$\pm$0.41 & 92.24$\pm$0.32 \\
        \hline
        \multirow{4}{*}{CCC (Ours)}   & Best & \textbf{80.77$\pm$0.10} & \textbf{81.84$\pm$0.14} & \textbf{83.32$\pm$0.13} & \textbf{82.72$\pm$0.11} & \textbf{91.93$\pm$0.09} & \textbf{92.87$\pm$0.06} & \textbf{93.01$\pm$0.07} & \textbf{93.18$\pm$0.10} \\
                                      & M.I. & $\uparrow$ 6.31 & $\uparrow$ 3.71 & $\uparrow$ 3.30 & $\uparrow$ 1.73 & $\uparrow$ 1.10 & $\uparrow$ 1.35 & $\uparrow$ 1.07 & $\uparrow$ 0.87\\\cline{3-10}
                                      & Last & \textbf{80.07$\pm$0.22} & \textbf{81.50$\pm$0.09} & \textbf{83.07$\pm$0.13} & \textbf{82.53$\pm$0.20} & \textbf{91.75$\pm$0.11} & \textbf{92.70$\pm$0.07} & \textbf{92.88$\pm$0.07} & \textbf{93.10$\pm$0.09}  \\
                                      & M.I. & $\uparrow$ 7.23 & $\uparrow$ 3.92 & $\uparrow$ 3.70 & $\uparrow$ 1.68 & $\uparrow$ 6.43 & $\uparrow$ 4.10 & $\uparrow$ 1.06 & $\uparrow$ 0.86\\
        \hline
    \end{tabular}%
    \caption{Comparison with state-of-the-art approaches in test accuracy ($\%$) on synthetic datasets with independent (IND) and correlated (COR) confusions. The best results are in bold. The ``M.I.'' represents minimal improvement.}
    \label{syn-ind}
\end{table*}

\section{Experiments}

We conduct experiments on two types of synthetic datasets (independent and correlated confusion) and three real-world datasets. To make the results reliable, all experiments are repeated three times with different random seeds, both the best accuracy on test set during training and the accuracy at the last epoch is reported. All experiments are conducted on a single NVIDIA RTX 3090 GPU. Due to space limitation, we only present the brief descriptions of datasets and results in the main paper, more information about datasets and the implementation details can be found in the Appendix.

\myPara{Baselines.}~We compare our method with extensive state-of-the-art approaches. The compared approaches include: (1) Majority Vote~\cite{majorityvote}. (2) Crowd Layer~\cite{crowdlayer}. (3) Doctor Net~\cite{guan2018said}. (4) Max-MIG~\cite{maxmig}. (5) CoNAL~\cite{conal}. (6) UnionNet~\cite{unionnet}.


\subsection{Evaluation on Synthetic Datasets}\label{syntheticdatas}
\myPara{Datasets.}~We evaluate our method on a variety of synthetic datasets, which are derived from two widely used datasets, i.e., {CIFAR-10}~\cite{cifar} and {Fashion-MNIST}~\cite{fmnist}. In general, we divide the synthetic datasets into two types: 1) \textit{independent confusion}, where the confusions of annotators are independent from each other and 2) \textit{correlated confusion}, where the confusions among different annotators may affect each other. Due to the limited page, we provide the generation details of these two confusions in the Appendix.

\myPara{Results.}~Table~\ref{syn-ind} shows the results on synthetic datasets with independent and correlated confusions. As shown, our CCC achieves superior performance than state-of-the-art methods across all scenarios. Based on the minimal improvement (M.I.) results, it is evident that the M.I. achieved at the last epoch is notable, which suggests that our CCC effectively prevents over-fitting while improving the model's best performance. On the other hand, the common baseline approach {Majority Voting} performs worst consistently, and the accuracy at last epoch decreases a lot compared to the best one, which is consistent with other works~\cite{maxmig,conal,tcl,crowdlayer,cvl}. This is because the naive majority vote fails to consider the difference among confusions of annotators, making the DNN eventually over-fits to the noisy labels. As for the approaches that consider annotators differently, their results are unstable, none of them performs better than any other methods consistently. The accuracy drop is also observed between the best and last epochs, the reason of which lies within that the ASCPs (e.g. confusion matrices in \cite{crowdlayer,maxmig}) of some annotators are poorly modeled, which makes the model finally memorize the noisy labels. It is worth noting that without using meta set to correct the learned confusion matrices, CCC will reduce to {CrowdLayer}~\cite{crowdlayer}. With the use of the meta set, the results of our CCC is significantly improved compared to that of {CrowdLayer}~\cite{crowdlayer}. This suggests that the confusion matrices employed in CCC are better able to model the expertise of annotators.

\subsection{Evaluation on Real-World Datasets}\label{exp_real}
\myPara{Datasets.}~Three real-world datasets are adopted for evaluation, including {LabelMe}~\cite{crowdlayer}, {CIFAR-10N}~\cite{cifarn} and {MUSIC}~\cite{music}. The first two datasets are image classification datasets, while the last one is a music genre classification dataset. The annotations of these real-world datasets are collected on \textit{Amazon Mechanical Turk}.
\begin{table}[tb]
    \centering
    \small
    \setlength\tabcolsep{4.5pt}
    \begin{tabular}{ll|c|c|c}
        \hline
        \multicolumn{2}{l|}{Method / Dataset} & LabelMe & CIFAR-10N & MUSIC \\
        \hline
        \multirow{2}{*}{MajorVote}   & Best & 80.72${\pm 0.45}$ & 81.08${\pm 0.34}$ & 67.67$\pm$0.98 \\
                                     & Last & 77.19${\pm 1.18}$ & 80.89${\pm 0.25}$ & 62.00$\pm$1.05 \\
        \hline
        \multirow{2}{*}{CrowdLayer}   & Best & 84.01${\pm 0.36}$ & 80.43${\pm 0.25}$ & 71.67$\pm$0.50 \\
                                      & Last & 82.32${\pm 0.41}$ & 80.14${\pm 0.28}$ & 69.33$\pm$0.93 \\
        \hline
        \multirow{2}{*}{DoctorNet}   & Best & 82.32${\pm 0.49}$ & 83.68${\pm 0.33}$ & 67.33$\pm$0.74 \\
                                     & Last & 81.73${\pm 0.49}$ & 83.32${\pm 0.29}$ & 65.33$\pm$0.75 \\
        \hline
        \multirow{2}{*}{Max-MIG}   & Best & 86.28${\pm 0.35}$ & 83.25${\pm 0.47}$ & 75.33$\pm$0.69 \\
                                   & Last & 83.16${\pm 0.66}$ & 83.12${\pm 0.63}$ & 71.67$\pm$0.93 \\
        \hline
        \multirow{2}{*}{CoNAL}      & Best & 87.46${\pm 0.53}$ & 82.51${\pm 0.15}$ & 74.00$\pm$0.89 \\
                                    & Last & \textbf{84.85${\pm}$0.91} & 80.92${\pm 0.28}$ & 68.67$\pm$1.88 \\
        \hline
        \multirow{2}{*}{UnionNet}   & Best & 85.19${\pm 0.42}$ & 82.66${\pm 1.12}$ & 72.33$\pm$0.58 \\
                                    & Last & 82.66${\pm 0.38}$ & 82.32${\pm 1.09}$ & 68.33$\pm$0.91 \\
        \hline
        \multirow{4}{*}{CCC (Ours)}   & Best & \textbf{87.65$\pm$1.10} & \textbf{86.36$\pm$0.05} & \textbf{76.22$\pm$0.42}\\
                                      & M.I. & $\uparrow$ 0.19 & $\uparrow$ 2.68 & $\uparrow$ 0.89 \\\cline{3-5}
                                      & Last & 84.79$\pm$0.80 & \textbf{86.12$\pm$0.12} & \textbf{72.89$\pm$0.68}\\
                                      & M.I. & $\downarrow$ 0.06 & $\uparrow$ 2.80 & $\uparrow$ 1.22 \\
        \hline
    \end{tabular}%
    \caption{Comparison with state-of-the-art approaches in test accuracy ($\%$) on real-world datasets. The best results are in bold. The ``M.I.'' represents minimal improvement.}
    \label{real_world}
\end{table}
\myPara{Results.}~The evaluation results on three real-world datasets are presented in Table~\ref{real_world}. As shown, our CCC outperforms other methods in two metrics (best and last) on {CIFAR-10N} and {MUSIC}. Additionally, on the {LabelMe} dataset, our CCC yields a new state-of-the-art result in terms of the best accuracy during training and achieves competitive accuracy results at the last epoch. These results provide evidence that our CCC is effective in handling real-world crowd-sourced annotations. Remarkably, the results of our CCC on {CIFAR-10N} are even comparable to the results that were trained with clean labels (refer to the Appendix). This means that the expertise of annotators in the {CIFAR-10N} dataset has been fully captured by our CCC.

\subsection{Ablation Study}\label{ablation}
\begin{figure}[tb]
\centering
    \includegraphics[width=0.42\textwidth]{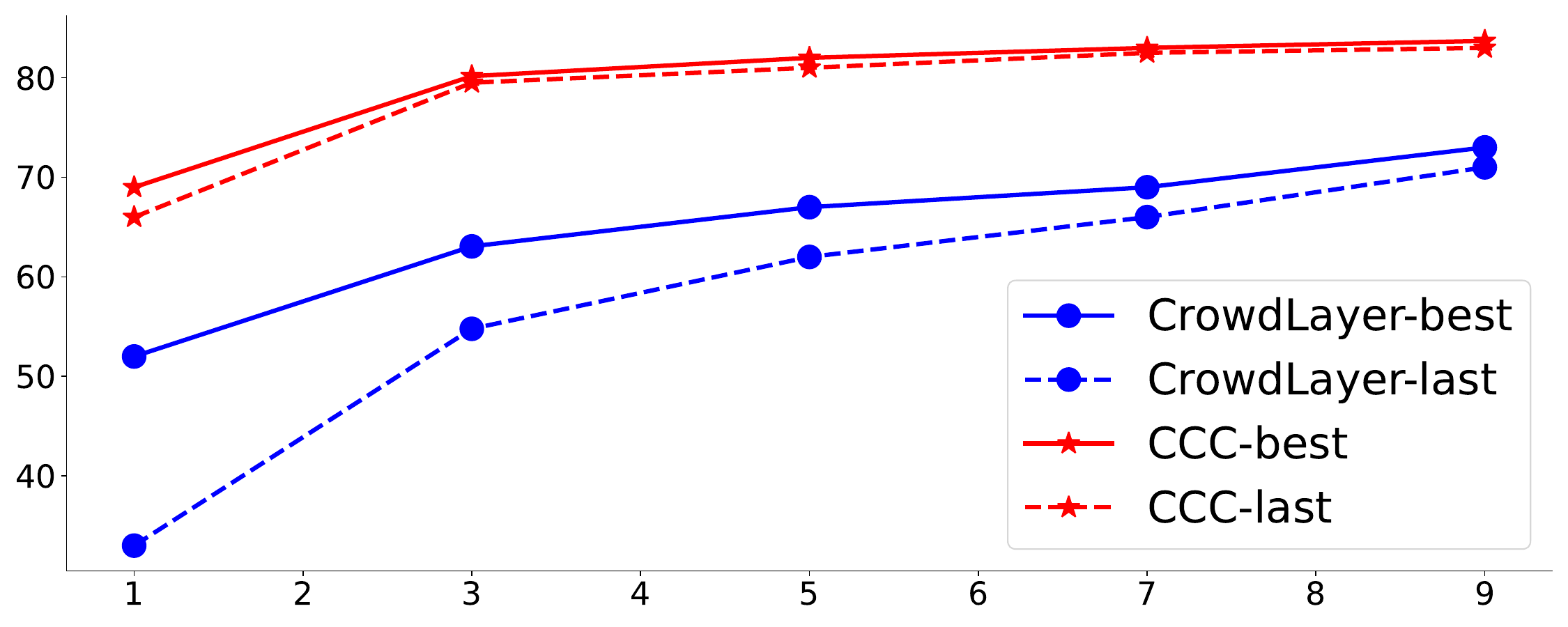}\\
    \includegraphics[width=0.42\textwidth]{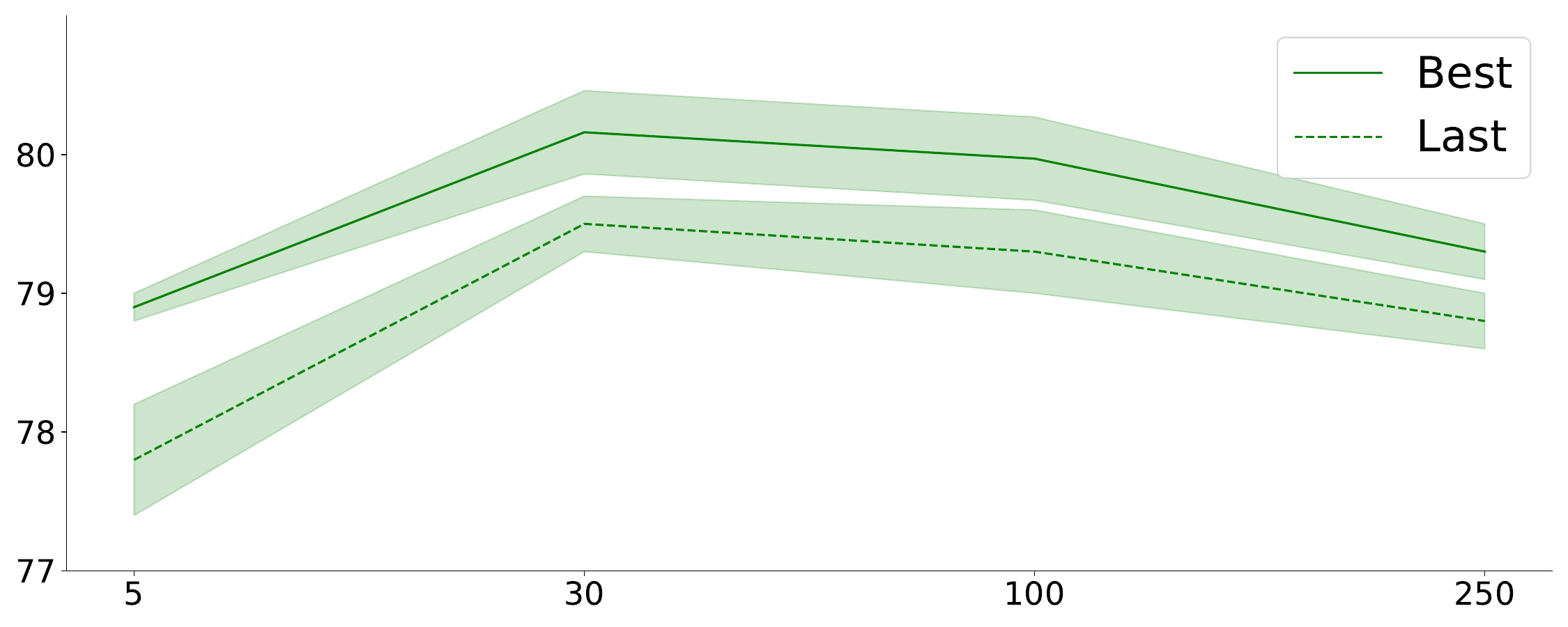}
\caption{The ablation studies on sparsity level (top) and number of annotator groups (bottom). The horizontal axis represents average number of annotators (top) and number of annotator groups (bottom), while the vertical axis denotes test accuracy.}
\label{ablation_plot}
\end{figure}

\myPara{Annotation Sparsity.}~To study the performance of CCC under different sparsity levels, we conduct experiments under different average numbers of annotations. To this end, we synthesize 4 more sets of annotations for the {CIFAR-10}~\cite{cifar} benchmark with an average number of annotations of 1,5,7,9, the confusion patterns of these datasets are identical to {CIFAR-10-IND-\uppercase\expandafter{\romannumeral1}}. The first chart in Figure~\ref{ablation_plot} depicts the corresponding results compared to {CrowdLayer}~\cite{crowdlayer}, from which we can see that our CCC yields higher test accuracy in both best and last metrics. Remarkably, even in the sparsest case, our CCC can still bring significant improvement (17.15$\%$ on best and 32.95$\%$ on last).

\myPara{Number of Annotator Groups.}~As shown in Figure~\ref{group}, there are various annotator groups in crowd-sourcing datasets, hence it is not necessary to consider all the annotators separately. However, the number of annotator groups is hard to identify, because the true labels are not available in practice. Therefore, in our CCC, the number of annotator groups $G$ is deemed as a hyper-parameter. We conduct the ablation study of $G$ on the {CIFAR-10-IND-\uppercase\expandafter{\romannumeral1}} dataset. The second chart in Figure~\ref{ablation_plot} demonstrates the test accuracy ($\%$) over different $G$'s. If $G$ is small, distinct annotators may be grouped together, leading to potential interference between their annotations. Conversely, if $G$ is large, similar annotators may be considered separately, resulting in poor modeling of their expertise due to annotation sparsity. 

\section{Conclusions}
In this work, we proposed a novel framework called {\textbf{C}}oupled {\textbf{C}}onfusion {\textbf{C}}orrection (CCC) to learn from multiple annotators, where the confusion matrices learned by one model can be corrected by the meta set distilled from the other. To further alleviate annotation sparsity, we unite various annotators based on their expertise using {K-Means} so that their confusion matrices could be corrected together. Besides, we reveal a common but neglected phenomenon in crowd-sourcing datasets that there are always annotators who provide seldom labels, which hinders the modeling of their expertise. Through extensive experiments across both synthetic and real-world datasets, we show that our CCC significantly outperforms other state-of-the-art approaches.


\section*{Acknowledgments}
This work was partially supported by grants from the National Key Research and Development Plan (2020AAA0140001), Beijing Natural Science Foundation (19L2040), and Open Research Project of National Key Laboratory of Science and Technology on Space-Born Intelligent Information Processing (TJ-02-22-01).

\nocite{*}
\bibliography{aaai24}

\clearpage
\appendix

\section{Related Works}
The problem we focused on in this paper is learning from crowds, and the learning paradigm we used is meta learning. Therefore, we hereby briefly review the related works from these two disciplines:

\myPara{Learning from Crowds (LFC).}~In the crowd-sourcing setting, one instance is associated with multiple labels, some of which could be noisy. To mitigate the influence of label noise, one common direction is to aggregate such noisy labels into a more reliable one and train the downstream tasks with these aggregated labels. For example, the {Majority Voting}~\cite{majorityvote,han2019beyond,zhou2012ensemble} is one of the most simplest and widely-used methods in aggregating crowd-sourcing labels. It naively assumes that all the annotators are equally reliable (i.e. have the same expertise), which is criticized for not able to handle the scenario where most annotators have relatively low expertise~\cite{weightedmajor, maxmargin, tensorfactor-1,tensorfactor-2}. Further, some more advanced aggregated methods are proposed, where the expertise of annotators are deemed to be different, including the {Weighted Majority Voting}~\cite{weightedmajor}, max-margin majority voting~\cite{maxmargin}, tensor factorization approaches~\cite{tensorfactor-1,tensorfactor-2}. Although {Majority Voting} and its variants provide a way to alleviate label noises, it is not always feasible to aggregate the labels, and these pipeline of methods may suffer from error propagation~\cite{majorityvote,han2019beyond,zhou2012ensemble}. 

On the other hand, by viewing the true label as an un-observable latent variable, many approaches tend to infer the true labels based on the {\textbf{E}}xpectation–{\textbf{M}}aximization (EM) algorithm~\cite{dawid1979maximum}. Among them, the Dawid and Skene's model~\cite{dawid1979maximum} serves as the basis of many works for aggregating labels from different annotators with different levels of expertise. For example, the approach proposed in~\cite{majorityvote} extends Dawid and Skene's model by jointly accounting for item difficulty. Similarly, \cite{ipeirotis2010quality} proposed to extract a confident score using Dawid and Skene's model, where the extracted score can be further used to prune low-quality annotators.

Despite the effectiveness of these EM-based algorithms, the training process often suffers from its computational complexity. This issue can be more severe when DNNs are adopted as the classifier, for the DNNs usually have massive parameters. To this end, many approaches are designed to directly model the annotator-specific expertise so as to use all the annotations on the fly~\cite{aggnet,conal,maxmig,tcl,crowdlayer}. For instance, the {CrowdLayer}~\cite{crowdlayer} is proposed to train a DNN directly with crowd-sourcing labels. Multiple ``crowd layers'' (linear layers without biases, also called confusion matrices) are added to the last softmax layer of the backbone DNN to capture the annotator-specific expertise of each annotator~\cite{crowdlayer}, in which way the backbone DNN and the ``crowd layers'' can be optimized simultaneously during training. Following this line of work, {CoNAL}~\cite{conal} proposed an approach to model a common confusion matrix except for the individual ones. {CoNAL} assumes that the annotation noise is attributed to two sources (common noise and individual noise), and combines these two noise by a Bernoulli random variable. Moreover, {UnionNet}~\cite{unionnet} unites all the noisy labels into a single ``united label'' in an one-hot manner and model the confusions by a single confusion matrix. In addition, there are some methods that do not rely on the confusion matrices but model the expertise of each annotator respectively as the {CrowdLayer} and {CoNAL} do. For example, the {DoctorNet}~\cite{guan2018said} aims to train a DNN with multiple softmax outputs to exploit all the annotations. Compared with the aforementioned aggregated methods like {Majority Voting} and its variants, the algorithms that learn the ASCPs (confusion matrices or multiple softmax outputs) simultaneously have achieved equal or even more promising performance. However, one of the major limitations of these methods is that learning such ASCPs usually suffers from the annotation sparsity, which is elaborated in Section~2.2.
\begin{table*}[tb]
    \centering
    \resizebox{\textwidth}{!}{
    \begin{tabular}{l|l|l|ccc|c|c|c|c|c|c|c|c|c|c|c|c}
        \hline
        \multicolumn{3}{c|}{\multirow{2}{*}{Dataset}}  & \multicolumn{3}{c|}{Split} & N.R.$_1$ & N.R.$_2$ & \multicolumn{2}{c|}{$\#$Annotators} & \multicolumn{8}{c}{Experiment Configuration} \\
        \cline{4-6} \cline{9-18}
        \multicolumn{3}{c|}{}     & $\#$Train &$\#$Test &$\#$Valid               & $(\%)$ & $(\%)$ & Total & Mean & Arch. & B.S. & Optim. & I.L.R. & \multicolumn{2}{c|}{$\#$Epochs} & $M$ & $G$\\
        \hline
            \multirow{9}{*}{\rotatebox{90}{ - CIFAR - }} &  \multicolumn{2}{c|}{10N} & \multirow{9}{*}{45K} & \multirow{9}{*}{10K} & \multirow{9}{*}{5K} & 02.13 & 17.71 & 747 & \multirow{17}{*}{3}    & \multirow{9}{*}{\rotatebox{90}{ResNet-34}} & \multirow{19}{*}{128} & \multirow{17}{*}{SGD} & \multirow{17}{*}{10$^{-2}$} & \multirow{17}{*}{60} & \multirow{17}{*}{10} & \multirow{18}{*}{1K} & 50\\
            \cline{2-3} \cline{9-9} \cline{18-18}
       & \multirow{4}{*}{\rotatebox{90}{IND}} & \uppercase\expandafter{\romannumeral1} &&&& 19.61 & 56.88 &  \multirow{16}{*}{250} &  &  &  &  & & & & & \multirow{16}{*}{30} \\
       & & \uppercase\expandafter{\romannumeral2} &&&& 26.87 & 63.44 &      &  &  &  &  & & & & &\\
       & & \uppercase\expandafter{\romannumeral3} &&&& 20.75 & 56.75&     &  &  &  &  &  & & & &\\
       & & \uppercase\expandafter{\romannumeral4} &&&& 09.68 & 46.21 &        &  &  &  &  &  & & & &\\
       \cline{2-3}
       & \multirow{4}{*}{\rotatebox{90}{COR}}& \uppercase\expandafter{\romannumeral1} &&&& 17.02 & 54.12 &        &  &  &  &  &  & & & &\\
       & & \uppercase\expandafter{\romannumeral2} &&&& 14.18 & 50.70 &        &  &  &  &  &  & & & &\\
       & & \uppercase\expandafter{\romannumeral3} &&&& 09.08 & 44.99 &        &  &  &  &  & & & & &\\
       & & \uppercase\expandafter{\romannumeral4} &&&& 10.07 & 46.72 &        &  &  &  &  &  & & & &\\
        \cline{1-6} \cline{11-11}
        \multirow{8}{*}{\rotatebox{90}{ - Fashion-MNIST - }} & \multirow{4}{*}{\rotatebox{90}{IND}}
           & \uppercase\expandafter{\romannumeral1} & \multirow{8}{*}{54K} & \multirow{8}{*}{10K} & \multirow{8}{*}{6K} & 18.73 & 55.82 &  &  & \multirow{8}{*}{\rotatebox{90}{ResNet-18}} &  &  &  &  & & &\\
       & & \uppercase\expandafter{\romannumeral2} &&&& 26.14 & 62.62&      &  & & &  &  & & & &\\
       & & \uppercase\expandafter{\romannumeral3} &&&& 21.45 & 57.98 &     &  &  &  &  & & & & &\\
       & & \uppercase\expandafter{\romannumeral4} &&&& 09.66 & 45.97 &     &  &  &  &  &  & && &\\
       \cline{2-3}
       & \multirow{4}{*}{\rotatebox{90}{COR}} & \uppercase\expandafter{\romannumeral1} &&&& 19.39 & 56.64 &     &  &  &  &  & & & & &\\
       & & \uppercase\expandafter{\romannumeral2} &&&& 15.77 & 51.74 &     &  &  &  &  & & & & &\\
       & & \uppercase\expandafter{\romannumeral3} &&&& 09.27 & 45.03 &     &  &  &  &  & & & & &\\
       & & \uppercase\expandafter{\romannumeral4} &&&& 10.67 & 47.66 &     &  &  &  &  & & & & &\\
        \cline{1-6} \cline{9-11} \cline{13-16} \cline{18-18}
        \multicolumn{3}{c|}{LabelMe}   & 10K & 1188  & 500 & 11.10 & 25.95 & 59 & 2.5  & VGG-16 &  & \multirow{2}{*}{Adam} & 10$^{-4}$ & \multirow{2}{*}{200} & \multirow{2}{*}{50} &  &25\\
        \cline{17-18}
        \multicolumn{3}{c|}{MUSIC}   & 700 & 300  & 0 & 12.71 & 43.97 & 44 & 4.2  & FC &  &  & 10$^{-2}$ &  &  & 160 &20\\
        \hline
    \end{tabular}
    }
        \caption{The summary of datasets information and experiment configurations. The first and second columns of $\#$Epochs is the total epochs for training and warm-up epochs respectively. Abbreviations: Arch.: the architecture of the backbone network; B.S.: batch size; Optim.: the type of optimizer; I.L.R.: initial learning rate; $M$: the size of meta set; $G$: the number of annotator groups.}
    \label{overall_info}
\end{table*}

\myPara{Meta Learning.}~Inspired by the developments of meta-learning\cite{metaapply_1,metaapply_2,metaapply_3}, many approaches are proposed to train a DNN that are robust to label noise~\cite{mwnet,mlc,mlc_microsoft,famus}. For instance, {MLC}~\cite{mlc} designed a loss correction method where the transition matrix is learned via meta-learning. And the transition matrix is further used to correct the loss during training, making the empirical risk over noisy dataset to be consistent with the risk over clean dataset. \cite{mlc_microsoft} use meta-learning to train a label-correction model during training, where the true labels can be inferred given instance features and noisy labels. In addition, meta-learning is also applied to learn an adaptive weighting scheme in~\cite{mwnet}, through which the weight of instances are adjusted. However, the great capability of meta learning has not been exploit in crowd-sourcing setting yet, which also motivates us to explore its performance in LFC.

\section{Pseudo Code of CCC}\label{pseudocode}
The pseudo code of CCC is presented in Algorithm~\ref{pseudocode_ccc}
\begin{algorithm*}[tb]
    \caption{Coupled Confusion Correction (CCC)}
    \label{pseudocode_ccc}
    \textbf{Input}: main parameters $\bm{\theta}_1, \bm{\theta}_2, \{\bm{T}^r_1, \bm{T}^r_2\}_{r=1}^R$, meta parameters $\{\bm{V}^r_1, \bm{V}^r_2\}_{r=1}^G$, training set $\tilde{\mathcal{D}}$, max epochs $E$, warmup epochs $E_w$
    \begin{algorithmic}[1] 
    \FOR{$e=0,...,E-1$}
        \IF {$e<E_w$}
        \STATE $\bm{\theta}_1, \bm{\theta}_2 = $ Warmup $(\tilde{\mathcal{D}}, \bm{\theta}_1, \bm{\theta}_2)$ //\textcolor{blue}{\textit{warmup the model using CrowdLayer to prepare for the distillation stage}}
        \ELSIF{$e<E$}
        \STATE $\mathcal{D}^{meta}_1 \stackrel{\text{distill}}{\underset{\bm{\theta}_2}{\longleftarrow}} \tilde{\mathcal{D}},\quad \mathcal{D}^{meta}_2 \stackrel{\text{distill}}{\underset{\bm{\theta}_1}{\longleftarrow}} \tilde{\mathcal{D}}$ //\textcolor{blue}{\textit{coupled distillation}}
        \STATE Initialize $\{\bm{V}^r_1, \bm{V}^r_2\}_{r=1}^G$ as zero tensors
        \STATE Get the group index of each annotator using {K-Means} \\$g(r) = \text{K-Means}(\{\bm{T}^r_1, \bm{T}^r_2\}_{r=1}^R)$
        \FOR{$k=1,2$}
            \FOR{$i=1$ to \textit{num}$\_$\textit{iters}}
            \STATE Fetch training-batch $\tilde{\mathcal{M}}\gets \tilde{\mathcal{D}}$
            \STATE Fetch meta-batch $\mathcal{M}^{meta}_k\gets \mathcal{D}^{meta}_k$
            \STATE Update $\{\bm{V}^r_k\}_{r=1}^G$ according to Eq.~(\ref{virtual}) and Eq.~(\ref{meta})\\
            //\textcolor{blue}{\textit{outer loop of the bi-level optimization}}
            \STATE Update $\bm{\theta}_k, \{\bm{T}^r_k\}_{r=1}^R$ according to Eq.~(\ref{actual})\\
            //\textcolor{blue}{\textit{inner loop of the bi-level optimization}}
            \ENDFOR
        \ENDFOR
        \ENDIF
        \ENDFOR
        \STATE \textbf{return} $\bm{\theta}_1, \bm{\theta}_2$
    \end{algorithmic}
\end{algorithm*}

\begin{table*}[tb]
    \centering
    \begin{tabular}{l|cccc|cccc}
    \hline
     Confusion Type & \multicolumn{4}{c|}{Independent Confusion} & \multicolumn{4}{c}{Correlated Confusion}\\ \hline
     Set Index & IND-\uppercase\expandafter{\romannumeral1} & IND-\uppercase\expandafter{\romannumeral2} & IND-\uppercase\expandafter{\romannumeral3} & IND-\uppercase\expandafter{\romannumeral4} & COR-\uppercase\expandafter{\romannumeral1} & COR-\uppercase\expandafter{\romannumeral2} & COR-\uppercase\expandafter{\romannumeral3} & COR-\uppercase\expandafter{\romannumeral4} \\\hline
     \multirow{2}{*}{{Pattern-\uppercase\expandafter{\romannumeral1}}} & {sym.} & {sym.} & {pair} & {sym.} & {sym.} & {pair} & {pair} & {sym.}\\
      & 0.3 & 0.4 & 0.3 & 0.3 & 0.4 & 0.5 & 0.4 & 0.5\\\hline
     \multirow{2}{*}{{Pattern-\uppercase\expandafter{\romannumeral2}}} & {sym.} & {class} & {pair} & {sym.} & {class} & {class} & {sym.} & {pair}\\
     & 0.5 & 2,5,9 & 0.6 & 0.5 & 2,5,9 & 0,6,8 & 0.5 & 0.7\\\hline
     \multirow{2}{*}{{Pattern-\uppercase\expandafter{\romannumeral3}}} & {pair} & {pair} & {class} & {sym.} & \multirow{2}{*}{{dummy}} & \multirow{2}{*}{{supp.}} & \multirow{2}{*}{{oppo.}} & {pair}\\
      & 0.6 & 0.6 & 0,4,5 & 0.7 & &  &  & 0.3\\\hline
     \multirow{2}{*}{{Pattern-\uppercase\expandafter{\romannumeral4}}} & {class} & {class} & {class} & {pair} & \multirow{2}{*}{{supp.}} & \multirow{2}{*}{{oppo.}} & \multirow{2}{*}{{supp.}} & \multirow{2}{*}{{oppo.}}\\
     & 1,3,4,6,8 & 0,6,8 & 1,3,4,6,8 & 0.5 &  &  &  & \\\hline
     \multirow{2}{*}{{Pattern-\uppercase\expandafter{\romannumeral5}}} & \multirow{2}{*}{{dummy}} & \multirow{2}{*}{{dummy}} & \multirow{2}{*}{{dummy}} & {pair} & \multirow{2}{*}{{oppo.}} & \multirow{2}{*}{{copy}} & \multirow{2}{*}{{copy}} & \multirow{2}{*}{{supp.}}\\
     & &  &  & 0.3 & &  &  & \\
     \hline
    \end{tabular}
        \caption{The confusion patterns of the synthetic datasets. \textit{sym.}: symmetrical confusion; \textit{class}: class-wise confusion; \textit{supp.}: supportive confusion; \textit{oppo.}: opposite confusion.}
    \label{confusion_types}
\end{table*}

\section{More Experiment Details}\label{more_exp_details}

\subsection{Generation of Two confusions}
For \textbf{independent} confusion, we consider four types of confusion patterns to generate the crowd-sourcing annotations, including 1) \textit{symmetrical}-$\epsilon$: the annotator would provide true labels with probability $1-\epsilon$ and provide wrong labels uniformly with probability $\frac{\epsilon}{C-1}$, 2) \textit{pair}-$\epsilon$: the annotator may confuse with similar class-pairs, it would label correctly with probability $1-\epsilon$ and provide the label from another class with probability $\epsilon$, 3) \textit{class-wise}-$c_0,c_1,...c_w$: the annotator would label the instances from classes $c_0,c_1,...c_w$ correctly with probability $1$, but it will choose labels uniformly for instances from other classes, 4) \textit{dummy}: the annotator would choose labels uniformly with probability $\frac{1}{C}$ for all classes. Note that the \textit{dummy} type confusion simulates the case where some annotators pick labels randomly to get rewards on the crowd-sourcing platforms, which is also taken into consideration in~\cite{cifarn}. For \textbf{correlated} confusion, following~\cite{tcl}, three types of correlated patterns are included: 1) \textit{copy}, where the annotator randomly picks another annotator and copies its label, 2) \textit{supportive}, where the annotator gives the right answer if another randomly picked annotator's answer is right, and give uniformly distributed answers otherwise, 3) \textit{opposite}, which is on the contrary of supportive pattern, the annotator gives the right answer if another randomly picked annotator's answer is wrong, and give uniformly distributed answers otherwise. 

As stated before, in real-world crowd-sourcing datasets, there are always some annotators who only provide seldom labels (as shown in Figure~1). However, previous works failed to take that issue into consideration when synthesizing crowd-sourcing annotations, which is actually an ubiquitous and harmful one as we analysed in Section~2.2. Therefore, we propose to use Beta distribution when generating the crowd-sourcing annotations, the overall process can be summarized as follows:

\noindent$\bullet$ For each annotator, generate its annotations based on the true labels and its confusion pattern (pre-defined).

\noindent$\bullet$ For each instance, select $k$ annotators and prune the annotations of others, the probabilities of selecting each annotator is set to follow a Beta density function $P \sim Beta(\alpha,\beta)$.

In this way, the synthetic datasets will be more consistent with real-world ones. The parameters of Beta distribution is fixed as $\alpha=1.5, \beta=3$ and $k$ is set to 3 in all synthetic datasets. To cover more scenarios, for each type of confusion, we generate $4$ sets of synthetic annotations with independent/correlated confusion for each benchmark dataset. Each set has $5$ different confusion patterns, and $50$ annotators are generated according to each confusion pattern, with a total of $250$ annotators. Note that the confusion patterns are shared between the two benchmark datasets. The confusion patterns of both independent and correlated types are shown in Table~\ref{confusion_types} in the Appendix.

\subsection{Datasets Descriptions}\label{datadetail}

\myPara{Synthetic Datasets.}~{CIFAR-10} dataset~\cite{cifar} has $50K$ training images and $10K$ test images with shape $32\times32\times3$, each image is associated with a label from $10$ classes. {Fashion-MNIST}~\cite{fmnist} has $60K$ training images and $10K$ test images with shape $28\times28$, same as {CIFAR-10}, each image is associated with a label from $10$ classes. In our experiments on synthetic datasets, $10\%$ of the training set is retained as a validation set, and the rest training labels are manually flipped to noisy crowd-sourcing labels. The detailed confusion patterns of each confusion types are listed in Table~\ref{confusion_types}.

\myPara{Real-World Datasets.}~We use three real-world crowd-sourcing datasets in this paper, including {LabelMe}~\cite{crowdlayer}, {CIFAR-10N}~\cite{cifarn} and MUSIC~\cite{music}. The labels of the three real-world datasets were all collected from \textit{Amazon Mechanical Turk} by human. {LabelMe} is an $8$-class image classification dataset containing $2688$ images in total, including $1K$ training images, $1188$ test images and $500$ validation images. Its training images are labeled by 59 annotators in total, with an average of $2.5$ annotators per image. Following the image pre-processing approach in~\cite{crowdlayer,maxmig,conal}, the $1K$ training images are augmented to $10K$ images. {CIFAR-10N}~\cite{cifarn} dataset consists of the original {CIFAR-10} training images with re-collected labels. Each image in {CIFAR-10N} is labeled by $3$ annotators, with a total of $747$ annotators. {MUSIC} is a music genre classification dataset, which consist of $1K$ music pieces with 30 seconds length. The $1K$ music pieces are divided into a training set of $700$ pieces and a test set of $300$ pieces. All the music pieces are from 10 music genres and are labeled by 44 annotators in total, with an average of 4.2 annotators per piece.

The information of all the datasets we use is summarized in Table~\ref{overall_info}, where the noise rate (N.R.) is calculated as:
\begin{equation*}
    \begin{aligned}
        & N.R._1 = \frac{\sum_{i=1}^N \mathds{1}[y_i \notin \tilde{\bm{y}_i}]}{N}, \\
        & N.R._2 = \frac{\sum_{i=1}^N\sum_{j=1}^R \mathds{1}[\mathcal{A}_{i,j}=1] * \mathds{1}[\tilde{y}_i^r \neq y_i] }{\sum_{i=1}^N\sum_{j=1}^R \mathds{1}[\mathcal{A}_{i,j}=1]}.
    \end{aligned}
\end{equation*}
The $N.R._1$ metric represents the overall noise level of the dataset. Specifically, an instance's crowd-sourcing label $\tilde{\bm{y}}$ are not considered noisy as long as at least one annotator provides the correct label. In contrast, the $N.R._2$ metric represents the noise level of all the annotators: If an annotator provides a wrong label, its individual label $\tilde{y}$ will be considered noisy.

\subsection{Implementation Details}\label{detailbaseline}
All of our experiments are conducted on NVIDIA RTX3090 GPUs with Pytorch framework~\cite{paszke2019pytorch}. To ensure a fair comparison, we did not use any data augmentation techniques for all datasets except for the LabelMe dataset, as mentioned above. For all the experiments, the learning rate of \textit{virtual} and \textit{actual} stage are kept identical ($\eta_v=\eta_a$), while the learning rate of \textit{meta} stage $\eta_m$ is set automatically as following:
\begin{equation*}
    \eta_m = \gamma \cdot \frac{Max(\{\bm{T}^r\}_{r=1}^R)}{Max(\{|\bm{g}^{r}_{cor}|\}_{r=1}^R)},
\end{equation*}
where $\bm{T}^r$ refers to the confusion matrix of the $r$-th annotator, and $\bm{g}^r_{cor}$ refers to the gradient of its confusion-correcting matrix. $\gamma$ is termed as ``correction rate'', reflecting the extent to which the confusion matrices are corrected. 
Note that the ``learning rate'' will refer to both virtual and actual learning rates in the following. For the {CIFAR-10} and {Fashion-MNIST} datasets, we adopt a batch size of 128, an SGD optimizer with a momentum of 0.9, weight decay of 5e-4, and an initial learning rate of 0.01. The learning rate is divided by 10 after the 40-th epoch, 60 epochs are set in total, 10 of which are used for warming up. The ResNet-34 is used for {CIFAR-10} and ResNet-18 is used for {Fashion-MNIST}. The number of annotator groups $G$ is set to 30. For the {LabelMe} dataset, we adopt a batch size of 128, an Adam optimizer with a learning rate of 1e-4. 200 epochs are set in total, including 50 epochs for warming up and 150 for CCC. The number of annotator groups are set to be 25. Following~\cite{crowdlayer,tcl,unionnet}, we use a pre-trained VGG-16 network followed by a fullly-conneted (FC) layer with 128 units, ReLU activations and a softmax output layer. For {MUSIC}, the same FC layer and output layer as {LabelMe} is adopted, and batch normalization is performed. we use an Adam optimizer with a learning rate of 1e-2. The batch size is set to 128. Similar to {LabelMe}, 200 total epochs and 50 warming up epochs are set, but the number of annotator groups for {MUSIC} is set to 20. The correction rate $\gamma$ is set to 0.3 for {MUSIC} and 0.5 for the other datasets. Additionally, the size of the meta set $M$ is set to 160 for {MUSIC} and 1000 for the other datasets. It is worth noting that as evidenced by previous studies~\cite{famus,fast_reweighting}, the bi-level optimization is time-consuming due to the computation of hessian matrix, which makes the training inefficient. Following~\cite{fast_reweighting}, to accelerate the training process when computing the gradient of the meta loss for the {CIFAR-10} and {Fashion-MNIST}, we only activate the linear layer of the network and detach the other layers. All of the information of datasets, experiment configurations, and hyper-parameters are summarized in Table~\ref{overall_info}.

\subsection{Details of Baselines}
Many baseline approaches include the initialization of confusion matrices, for the Max-MIG~\cite{maxmig} and UnionNet~\cite{unionnet}, following their original paper, we adopt the initialization strategy as:
\begin{equation*}
    \bm{T}^r(p,q)=log\frac{\sum_{i=1}^N Q(y_i = p)\mathds{1}[\tilde{y}_i^r = q]}{\sum_{i=1}^N Q(y_i = p)},
\end{equation*}
in which the $Q(y_i = p)$ is obtained by averaging all crowd-sourcing labels:
\begin{equation*}
    Q(y_i = p) = \frac{1}{R}\sum_{r=1}^R\mathds{1}[\tilde{y}_i^r = p].
\end{equation*}
Moreover, the original paper of UnionNet~\cite{unionnet} proposed two variants of UnionNet, but we only take the UnionNet-B as baseline because the UnionNet-B performs better according to the results of its original paper~\cite{unionnet}. For CrowdLayer, following the settings in Max-MIG~\cite{maxmig} and its original paper~\cite{crowdlayer}, we initialize the confusion matrices as identity ones.

\begin{table}[h]
    \centering
    \resizebox{0.47\textwidth}{!}{
        \begin{tabular}{c|c|c|c|c}
        \hline
        Dataset & CIFAR-10 & F-MNIST & LabelMe & MUSIC\\ \hline
         Best & 86.19${\pm 0.28}$ & 93.80${\pm 0.15}$ & 91.84${\pm 0.10}$ & 82.00$\pm$1.14\\
         Last & 86.15${\pm 0.30}$ & 93.61${\pm 0.14}$ & 91.16${\pm 0.25}$ & 79.67$\pm$1.33 \\
        \hline
    \end{tabular}
    }
    \caption{Results that trained with clean (ground-truth) labels of all datasets we used. The ``F-MNIST'' represents the Fashion-MNIST dataset.}
    \label{results_gt}
\end{table}

\section{Results Trained with Clean Labels}\label{more_results}
The test accuracy ($\%$) trained on clean labels of all datasets we used in this paper is reported in Table~\ref{results_gt}.

\end{document}